\documentclass[10pt,twocolumn,letterpaper]{article}

\usepackage[pagenumbers]{cvpr} 

\usepackage[utf8]{inputenc} 
\usepackage[T1]{fontenc}    
\DeclareUnicodeCharacter{2212}{\ensuremath{-}}
\usepackage{url}            
\usepackage{booktabs}       
\usepackage{amsfonts}       
\usepackage{nicefrac}       
\usepackage{microtype}      
\usepackage{bbm}
\usepackage{amsmath}
\usepackage{amssymb}
\usepackage{enumitem}
\usepackage{bbding}
\usepackage{algorithm}
\usepackage{algpseudocode}
\usepackage{array}
\usepackage{listings}
\usepackage{multirow}
\usepackage{wrapfig}
\usepackage[dvipsnames,table,xcdraw]{xcolor}
\usepackage{graphicx}
\usepackage{caption}
\usepackage{tabularx}
\usepackage{pifont}
\usepackage{makecell}
\usepackage[margin=0.7in]{geometry} 

\newcommand{\addFig}[1]{}
\newcommand{\addFigs}[1]{}
\definecolor{cvprblue}{rgb}{0.21,0.49,0.74}
\usepackage[pagebackref,breaklinks,colorlinks,linkcolor=BrickRed,citecolor=cvprblue]{hyperref}

\graphicspath{{./figures/}}

\newtheorem{definition}{Definition}[section]

\newcommand{\Input}{\item[\textbf{Input:}]} 
\newcommand{\Output}{\item[\textbf{Output:}]} 

\begin{document}
\title{Contrastive Representation Learning for Dynamic Link Prediction \\ in Temporal Networks}

\author{Amirhossein Nouranizadeh $^*$ \quad Fatemeh Tabatabaei Far \thanks{Equal Contributions} \\ \quad Mohammad Rahmati \\ \\
Department of Computer Engineering \\
 Amirkabir University of Technology, Tehran, Iran\\
{\tt\small \{nouranizadeh, tabatabaeifateme, rahmati\}@aut.ac.ir}
}
\maketitle

\begin{abstract}
Evolving networks are complex data structures that emerge in a wide range of systems in science and engineering.
Learning expressive representations for such networks that encode their structural connectivity and temporal evolution is essential for downstream data analytics and machine learning applications.
In this study, we introduce a self-supervised method for learning representations of temporal networks and employ these representations in the dynamic link prediction task.
While temporal networks are typically characterized as a sequence of interactions over the continuous time domain, our study focuses on their discrete-time versions.
This enables us to balance the trade-off between computational complexity and precise modeling of the interactions.
We propose a recurrent message-passing neural network architecture for modeling the information flow over time-respecting paths of temporal networks.
The key feature of our method is the contrastive training objective of the model, which is a combination of three loss functions: link prediction, graph reconstruction, and contrastive predictive coding losses.
The contrastive predictive coding objective is implemented using infoNCE losses at both local and global scales of the input graphs.
We empirically show that the additional self-supervised losses enhance the training and improve the model's performance in the dynamic link prediction task.
The proposed method is tested on Enron, COLAB, and Facebook datasets and exhibits superior results compared to existing models. \footnote{\href{https://github.com/amrhssn/teneNCE}{https://github.com/amrhssn/teneNCE}}
\end{abstract}
\section{Introduction}\label{sec:intro}
Many processes in science and engineering are modeled as dynamical systems over evolving networks. 
These processes range from contagion processes over networks of individual contacts, like disease spreading, to information propagation over communication networks \cite{holme2012temporal}, such as those observed in social networks.
A key feature of processes supported on evolving networks is that their dynamics are highly coupled with the network’s topological and temporal features. This makes it essential to consider the evolving nature of networks when modeling such systems. 
\begin{figure}[t]
    \centering
    \includegraphics[width=0.95\columnwidth]{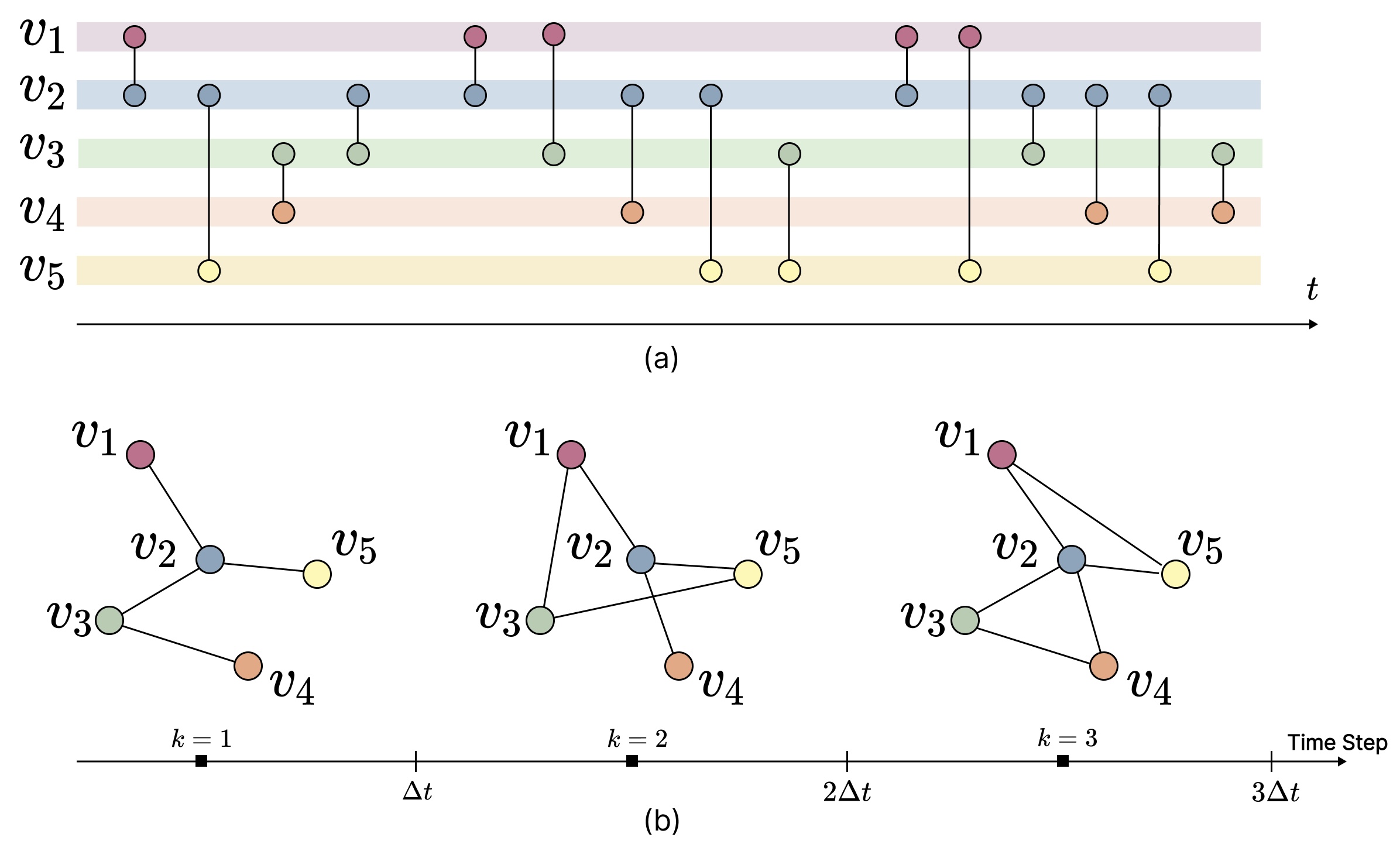} 
    \caption{(a) Illustration of a temporal network as a sequence of pairwise interactions between system entities over a continuous time interval. (b) The discrete-time snapshot sequence represents the temporal network. The discretization operation divides the temporal network’s time domain into equal-length intervals $\Delta t$ and projects the interactions within each time interval to a static graph.}
    \label{fig:temporal-network}
\end{figure}
In recent years, there have been numerous efforts in the machine learning community to encode the structures of graphs into real vector spaces, a field that is generally referred to as graph representation learning.
Graph neural networks (GNNs) constitute a significant paradigm in graph representation learning methods. 
In addition to the abundance and ubiquity of graph-based data and the availability of Graphical Processing Units (GPUs), the successful implementation of GNNs can be primarily attributed to the widespread accessibility of open-source software libraries, such as \cite{fey2019fast, wang2019dgl, tfgnn}.

However, most graph-based deep learning methods are typically applied to static graphs for representation learning. This contrasts with numerous real-world networks that are inherently dynamic. 
Nevertheless, there has been a growing interest in the development of methods that handle temporal networks and their deployment in production environments. 

In the present work, our goal is to learn vector representations of temporal networks in a manner that ensures the learned embeddings contain structural and temporal information of the underlying dynamic graph.
To this end, we model the evolution of the temporal network as a discrete-time dynamical system. 
The system is parameterized via a novel recurrent message-passing neural network architecture that naturally facilitates the flow of information over time-respecting paths of the dynamic graph. 

Moreover, to train the model's parameters, we formulate a loss objective that encourages the model to extract features of the temporal network that are informative for both transductive and inductive downstream machine learning tasks, namely the reconstruction and prediction of the dynamic graphs.
Additionally, to regularize the training, we include a self-supervised infoNCE loss that operates at the local and global scales of the temporal network and guides the model to encode features of the data that span longer periods into the future. 
Intuitively, this training strategy creates a balance between extracting low-level features that are informative for predicting the next state of the graph and high-level features that describe the graph far into the future. 

We call the proposed method \textit{\underline{Te}mporal \underline{Ne}twork \underline{N}oise \underline{C}ontrastive \underline{E}stimation}, \textit{teneNCE} for short. 
We evaluate the teneNCE representations in the dynamic link prediction task and numerically show that it achieves competitive results compared to state-of-the-art existing models.
Our main contributions are summarized as follows:
\begin{itemize}
    \item We introduce a novel recurrent message-passing neural network architecture for learning temporal representations in dynamic graphs. This model propagates node information through time-respecting paths in the temporal network.
    \item We propose a single loss function that integrates reconstruction, prediction, and self-supervised infoNCE losses.  The reconstruction loss helps the model in learning current structure of the temporal network, while the prediction loss ensures that the learned features are informative for predicting the network's structure in the next time step. Moreover, the infoNCE loss function effectively captures high-level dynamics that extend further into the future by operating at both the local node-level and the global graph-level of the temporal network. The training process demonstrates that infoNCE loss decreases continuously, enabling the model to extract more informative representations.
    \item The teneNCE model shows superior performance in the dynamic link prediction task on Enron, COLAB, and Facebook datasets, achieving an average improvement of 4\% compared to state-of-the-art models.
\end{itemize}
In the remainder of this paper, we will revisit the related works that have guided us in formulating our proposed method in Section \ref{sec:related_work}. Following this, we will elaborate on our method in Section \ref{sec:method} and subsequently discuss the experimental setup and numerical findings in Section \ref{sec:experiments}. Finally, we will conclude this work and outline our future research trajectory in Section \ref{sec:conc}.

\section{Related Works}\label{sec:related_work}
\subsection{GNNs for static graphs}
The field of graph representation learning has experienced significant success, mainly due to advancements in graph neural network research. 
These models are message-passing neural networks that encode the structure and features of a graph into an embedding space by exchanging neural messages across the topology of the graph \cite{gcn, graphsage, gilmer2017neural, gat, gae, gin, fastgcn}. 
Recent research in graph representation learning focuses on addressing the fundamental limitations of GNNs, such as oversmoothing and over-squashing problems \cite{oversmoothing, oversquashing1, oversquashing2}. 
Furthermore, to enhance the expressive power of GNNs, proposed methods have employed graph-rewiring techniques \cite{diffwire, oversquashing1}, graph transformers \cite{graphtransformer, san, shi2022benchmarking, ying2021transformers, mialon2021graphit, wu2021representing, graphgps}, and integration of topological inductive biases into these models to inform them about the underlying graph structure \cite{wang2022equivariant, lim2022sign, li2020distance, dwivedi2021graph, chen2022structure}.

\subsection{GNNs for dynamic graphs} 
Given the evolving nature of graphs in many real-world scenarios, dynamic graph representation learning is still in its early stages compared to static graph embedding methods.
However, there has been a recent interest in developing methods specifically tailored for temporal networks.

In general, deep learning models for dynamic graph representation learning can be classified into two categories, depending on how they represent the input data: discrete-time or continuous-time models.

\subsubsection{Discrete-time models}
Discrete-time models are applied to a sequence of static graph snapshots. 
GCRN \cite{gcrn} generalizes the convLSTM model \cite{convlstm} to graphs, where the 2D Euclidean convolutions in the RNN are replaced with graph convolutions. 
This modification allows the RNN model to perform message-passing operations over the graph structure to update its internal hidden state. 
However, in their original work, the GCRN model was used to model spatiotemporal sequences of images and text data, with the graph structure kept fixed during training. 

VGRNN \cite{vgrnn} extends the GCRN model to evolving graph sequences and is trained according to the variational autoencoder framework.  
More specifically, VGRNN introduces latent variables sampled from a prior Gaussian distribution, which is a function of previous graph structures that are captured by graph recurrent neural networks. 
These latent variables are then optimized to reconstruct future graphs.

In parallel to VGRNN, GC-LSTM \cite{gclstm} proposes an encoder-decoder architecture to learn the dynamic graph representations. 
The encoder model is a graph convolutional LSTM and is responsible for capturing temporal-topological information of the snapshot sequence, and the decoder model is a fully connected MLP that predicts the future structure of the graph.

EULER \cite{Euler} also adopts the standard architecture of integrating GNNs with sequential encoders for temporal link prediction.
Their proposed method aims to execute this architecture on large-scale graphs by introducing a parallel computational framework where separate machines run a replicated GNN.
This enables the architecture to process multiple snapshot graphs simultaneously, demonstrating the capability of scaling the sequential architecture in practice.

EvolveGCN \cite{evolvegcn} approaches the temporal graph embedding differently. 
Instead of training the model to encode the dynamics of graph structure into its parameters, EvolveGCN uses a recurrent neural network to update the parameters of the GNN model to react to the evolution of input graphs. 
This allows EvolveGCN to address the problem of frequently changing node sets in the dynamic graph.

While most discrete-time models employ RNNs to capture the time evolution of the snapshot sequence, DySAT \cite{dysat} uses self-attention layers to jointly encode the structural and temporal information of the dynamic graph. 
Through the use of positional embeddings, DySAT processes the sequence of snapshot graphs simultaneously, making it computationally more efficient compared to recurrent processing RNN-based methods.

\subsubsection{Continuous-time models}
Continuous-time models are directly applied to the temporal edge sequence observed over continuous timestamps.
These models use time encoding functions to embed the time dimension information into real vector spaces.
Time encoders can have both fixed \cite{graphmixer} and trainable parameters \cite{tgn, tgat}.
TGAT \cite{tgat} employs a self-attention mechanism to jointly capture the structural and temporal information of the temporal network. 
TGN \cite{tgn} initially captures each node's temporal information by aggregating the node's interaction information, which is represented as edge features, using a recurrent neural network.
Subsequently, TGN employs a graph attention neural network on the time-aggregated graph, enabling it to capture structural and temporal information.

In contrast to other dynamic GNN architectures that use RNNs or attention mechanisms for encoding temporal networks, GraphMIXER \cite{graphmixer} introduces an architecture composed solely of MLPs.
GraphMIXER uses a link encoder to compute edge information and then applies a node encoder to aggregate each node's messages over its links.

Interested readers are referred to \cite{kazemi2020representation} for a comprehensive survey on the dynamic graph representation learning task.

\subsection{Contrastive learning on graphs}
Contrastive learning (CL) is a popular self-supervised learning technique that learns representations by contrasting similar and dissimilar instances sampled from a data distribution.
CL has proved to be successful in computer vision \cite{chen2020simple, he2020momentum, grill2020bootstrap, han2019video}, natural language \cite{fang2020cert, gunel2020supervised, yan2021consert} and audio processing \cite{niizumi2021byol, saeed2021contrastive, spijkervet2021contrastive}.

Noise Contrastive Estimation (NCE) is introduced in \cite{nce} as a technique for estimating the parameters of unnormalized statistical models. 
The method trains a logistic regression classifier to discriminate data points from samples drawn from a known noise distribution.
Intuitively, as the authors articulated, the idea behind the noise-contrastive estimation is ``learning by comparison". 

In the well-known work \cite{oord2018representation}, the authors proposed the Contrastive Predictive Coding (CPC) technique for unsupervised representation learning of sequential data.
CPC encodes each element of the sequence via an encoder function. It then uses an autoregressive model to capture context representations that maximize mutual information with the future representations of the sequence elements. 
The mutual information maximization is done by optimizing the introduced infoNCE loss function.
In essence, the infoNCE loss can be interpreted as the binary cross entropy loss between the similarity scores of the context representation and the future encodings, contrasted with the similarity scores of the context representation and encodings of negative samples.
For further details on CPC and infoNCE loss, see Appendix \ref{subsec:infoNCE}.

Contrastive learning has also been shown to be an effective objective for capturing graph information through GNN architectures in unsupervised graph representation learning.
Deep Graph Infomax (DGI) \cite{dgi} extends Deep Infomax \cite{hjelm2018learning} to graph-structured data. 
The DIG model's learning objective is to differentiate between the similarity scores of local-global embedding pairs of the input graph and those derived from a shuffled version of the same graph.

Alternatively, InfoGraph \cite{infograph} employs deep Infomax to learn graph-level representations. It aims to maximize the mutual information between the graph-level representation and the representations of substructures at varying scales within the graph.
In the study \cite{hassani2020contrastive}, the authors construct a different structural view of the graph using the graph diffusion process and differentiate between the node and graph level representation of the two graph views to learn graph representations.

It is essential to mention that despite the success of contrastive learning methods in representation learning, it has been shown that the performance of these methods cannot solely be attributed to the properties of mutual information alone. 
It also depends on the choice of encoders and mutual information estimators \cite{tschannen2019mutual}. 
For a complete survey on graph representation learning with contrastive methods, refer to \cite{xie2022self, liu2022graph}.
\begin{figure*}[ht]
    \centering
    \includegraphics[width=0.95\textwidth]{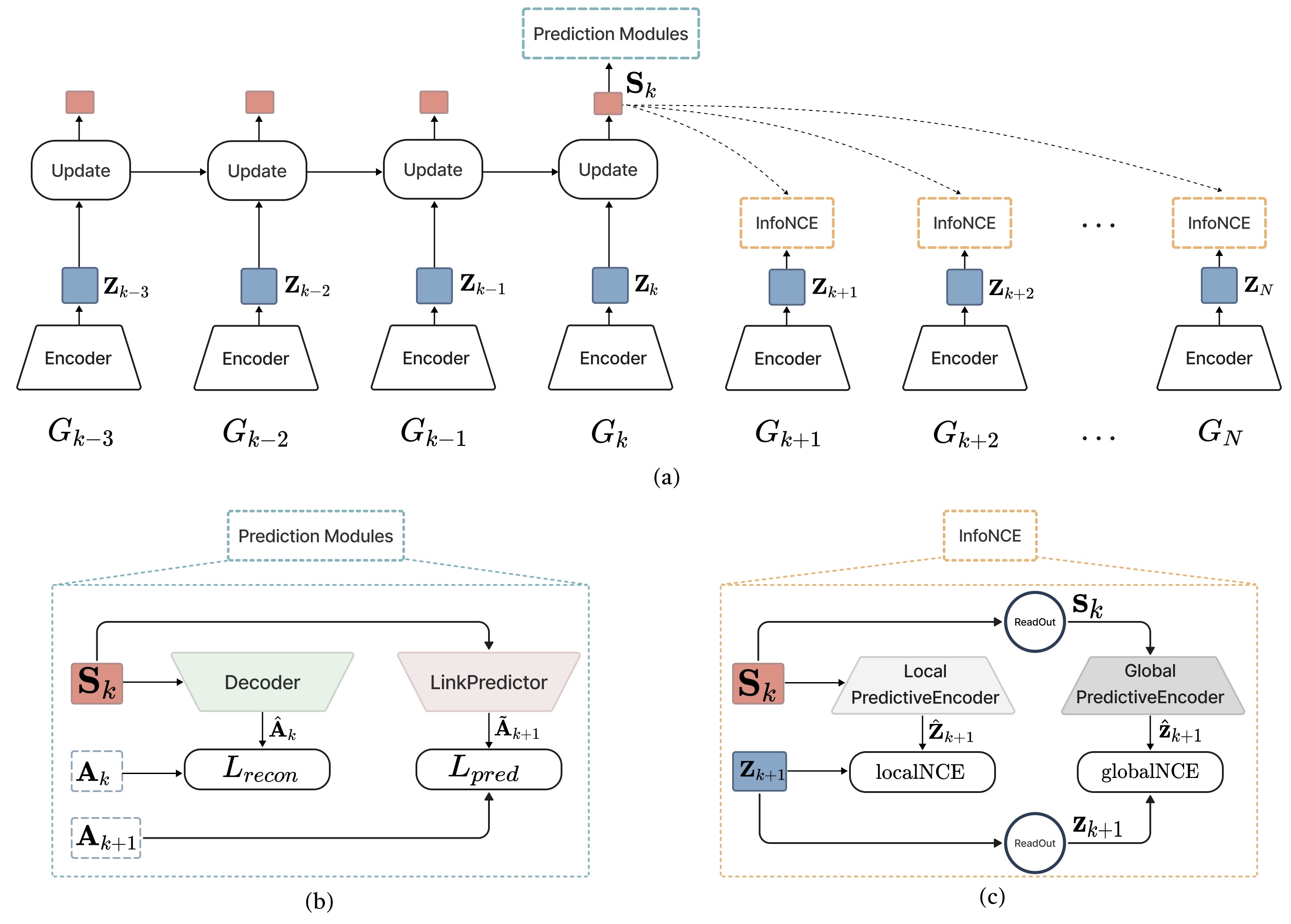} 
    \caption{
    (a) Illustration of the teneNCE model architecture. The model processes a sequence of snapshot graphs. The main components of the teneNCE model include an \textrm{Encoder} for embedding each static graph in the sequence and an \textrm{Update} component that recursively updates the state representations of each node across time steps. During the forward pass of the model in the training process, at time step $k$, the updated node states $\mathbf{S}_k$ are used to compute the (1) reconstruction loss, (2) prediction loss, and, (3) contrastive predictive coding loss.
    (b) The prediction modules include a \textrm{Decoder} for reconstructing the static graph at each time step and a \textrm{LinkPredictor} for predicting the graph's structure at the next time step. 
    (c) An overview of the CPC loss module, which consists of a \textrm{LocalPredictiveEncoder} and a \textrm{GlobalPredictiveEncoder} for predicting the future structural embeddings of the graph based on the node states, for time steps $k+1, \dots, N$.
    Additionally, the $\textrm{ReadOut}(.)$ function aggregates the node-level embeddings into the graph-level representation.}
    
    \label{fig:archicture}
\end{figure*}

\section{Methodology}\label{sec:method}
As briefly mentioned in Section \ref{sec:intro}, our goal is to learn node representations of a temporal network that capture the structural information of the network as well as its temporal evolution.
Specifically, we seek to learn a function that maps each node of the temporal network to an embedding vector, such that the embeddings contain sufficient information to predict the network's future structure.

The unique aspect of our approach in accomplishing this goal lies in the application of the infoNCE loss function at both local and global scales of the temporal network, which regulates the training process.
Intuitively, by incorporating both node-level and graph-level network evolution within the contrastive learning framework, we seek to regularize the model in learning node representations that align with the global network dynamics.
Our empirical evaluation demonstrates the efficacy of our approach in the dynamic link prediction task across three datasets: Enron, Colab, and Facebook. We find that it delivers competitive quantitative results when benchmarked against existing methods.

In the following subsections, we will detail the development process of the teneNCE model.
All the notations used throughout the paper are summarized in Table \ref{table:notations}.

\subsection{Preliminaries}
The main objects of our interest are temporal networks.
In most systems, where pairwise events occur or other elements within the system influence the dynamics of individual objects, the data is represented by a temporal graph. 
This data consists of an interaction sequence over time where interactions are events between system entities.
Since temporal networks represent the internal interactions of a system over time, additional time-dependent information can be attributed to them, such as node and edge features at the time of the interaction.
Specifically, we define temporal networks as follows:
\begin{definition}[Temporal network]\label{def-temporal-network}
	A temporal network $\mathcal{G}=(\mathcal{V}, \mathcal{E})$ is defined by the set of nodes $\mathcal{V}=\{v_1, \dots, v_n\}$ and the set of events $\mathcal{E}=\{e_{t_1}, \dots, e_{t_m}\}$, where $n$ and $m$ are the number of nodes and events, respectively. Each event $e_{t}=(v_i, v_j, t) \in \binom{\mathcal{V}}{2} \times \mathbb{R}_+$ represents a pairwise interaction between nodes $v_i$ and $v_j$ at the time $t \in \mathbb{R}_+$.
	In attributed networks, events and nodes can have extra time-dependent feature information, which can be represented by $e_{t}=(v_i, v_j, \mathbf{x}_i(t), \mathbf{x}_j(t), \mathbf{e}_{ij}(t), t)$ where $\mathbf{x}_i(t), \mathbf{x}_j(t) \in \mathbb{R}^{d_\mathcal{V}}$ and $\mathbf{e}_{ij}(t) \in \mathbb{R}^{d_\mathcal{E}}$ are the feature information of the nodes $v_i, v_j$ and event $e_{t}$ at time $t$, respectively.  Here $d_\mathcal{V}$ and $d_\mathcal{E}$ denote the dimensions of feature vectors for nodes and edges.
\end{definition}

Temporal networks are also referred to as \textit{continuous-time dynamic graphs} (CTDG) \cite{kazemi2020representation}.
Instead of processing continuous-time dynamic graphs, we work with discrete-time versions of temporal networks, also known as \textit{snapshot sequences}.
We project the temporal network into a sequence of static snapshot graphs, each capturing interactions within a specific, fixed time interval.
The discretization function divides the entire timespan of the temporal network into equally-distanced time intervals and projects the interactions of each interval into a time-aggregated static graph.
Despite losing some information during the discretization process, this approach allows us to handle a greater number of interactions in a single call to the representation model.
\begin{definition}[Snapshot sequence]
Given a temporal network $\mathcal{G}=(\mathcal{V}, \mathcal{E})$, a snapshot sequence $G=\{G_1, \dots, G_N\}$ is defined as the sequence of static graphs $G_k=(\mathcal{V}, E_k)$, where the set of edges is defined as $E_k=\{e_t \in \mathcal{E}: (k-1)\Delta t \leq t < k\Delta t \; \textrm{for} \; k=1, \dots, N, \Delta t=\frac{t_m - t_1}{N}\}$. The set of edges can be represented by the adjacency matrix $\mathbf{A}_k \in \mathbb{R}^{n \times n}$. Attributed snapshot graphs can be represented as $G_k=(\mathbf{X}_k, \mathbf{E}_k, \mathbf{A}_k)$, where $\mathbf{X}_k  \in \mathbb{R}^{n \times d_\mathcal{V}} $ and $\mathbf{E}_k \in \mathbb{R}^{|E_k| \times d_\mathcal{E}}$ are the feature matrices of nodes and edges, respectively.
\end{definition}

The snapshot sequence is also called a \textit{discrete-time dynamic graph} (DTDG) \cite{kazemi2020representation}.
We model the evolution of a temporal network as a discrete-time dynamical system that is parameterized by a composition of recurrent and graph neural networks.
After optimizing the model's parameters, the learned state representations of the nodes encode temporal-topological information of the system, which are computed as a function of historical data. 
These node state representations can be used in any downstream machine-learning task defined over the temporal network.

\subsection{Motivation}\label{subsec:motivation}
In a temporal network, information spreads through time-respecting paths. 
A time-respecting path is defined as a set of interactions, ordered by time, that connects subsets of nodes \cite{kempe2000connectivity} in a temporal network.
 For any given node $v_i$ at a specific time $t$, the set of all other nodes that have time-respecting paths to node $v_i$ up to time $t$, is called the \textit{source set} of $v_i$ \cite{holme2012temporal}. 
 Node states in a temporal network are influenced by their corresponding source sets.
 This makes the ideal modeling of a temporal network an inherently sequential process since the node states must be updated after each interaction. 

To model the information propagation over a temporal network, we need to consider the trade-off between updating node states after every interaction and the computational efficiency of our model.
This motivates us to work with the discretized version of the temporal network, which is a sequence of static graph snapshots, and concurrently update the node states of each static graph by a single forward pass of the model.
Furthermore, to compensate for the information loss resulting from the discretization process, we employ a message-passing mechanism on each static graph within the snapshot sequence before updating the node states.

To steer the model towards extracting state representations that reflect the temporal-topological information of the system, we define a learning objective that is constituted of three components:
\begin{enumerate}[label=\Roman*.]
    \item Initially, we want the node representations to contain the topological information of the network at any given time. This results in the formulation of the \textit{reconstruction loss}, which encourages the learned representations to contain the necessary information for reconstructing the graph at a given time.
    \item Next, we want the representations to capture the required temporal information for predicting the temporal network structure at the next time step. This motivates the \textit{prediction loss}.
    \item Finally, we want the representations to contain topological-temporal features of the network that span multiple time steps. To this end, we employ the \textit{contrastive predictive coding} loss function \cite{oord2018representation}, which maximizes the mutual information between node and graph level representations and corresponding future features of the temporal network.
\end{enumerate}

By optimizing the learning objective with respect to the model's parameters, we are able to embed the system's information into state representations and subsequently feed them into machine learning workflows of temporal networks.

\subsection{Model}
Our model consists of five components, as shown in Figure \ref{fig:archicture}.

\subsubsection{Encoder} 
The Encoder function maps the structural information of each snapshot graph into the structural embedding space. 
It is implemented using a graph neural network that acts on each static graph:
\begin{equation}
	\begin{aligned}
		\mathbf{Z}_k = \textrm{Encoder}(\mathbf{X}_k, \mathbf{A}_k) \quad k=1, \dots, N. \\
	\end{aligned}
\end{equation}

Here, $\mathbf{Z}_k \in \mathbb{R}^{n \times d_\textrm{enc}}$ is the structural embedding matrix, and $d_\textrm{enc}$ is the output dimension of the $\textrm{Encoder}$.
In practice, any GNN can be used as an encoder. 
In our experiments, we used a simple 3-layer graph convolutional network \cite{gcn} with $\textrm{ReLU}$ activation functions.

\subsubsection{Update} 
The function Update encodes the temporal network into the hidden state space by modeling time evolution through a discrete-time dynamical system, which generates state trajectories:
\begin{equation}\label{eq:update}
	\begin{aligned}
		& \mathbf{S}_{k} = \textrm{Update}(\mathbf{Z}_{k}, \mathbf{A}_{k}, \mathbf{S}_{k-1}, k) \quad k=1, \dots, N, \\
		& \mathbf{S}_0 = \mathbf{0}.
	\end{aligned}
\end{equation}
The state matrix $\mathbf{S}_{k} \in \mathbb{R}^{n \times d_\textrm{state}}$, which serves as node representations, is updated using the recurrent function $\textrm{Update}$, given the current structural embedding matrix $\mathbf{Z}_{k}$, current adjacency matrix $\mathbf{A}_{k}$, previous state matrix $\mathbf{S}_{k-1}$ and encoded current time step $k$ of the system. 
The dimension of the hidden state is denoted by$d_\textrm{state}$.

In our implementation, inspired by \cite{gcrn} and similar to \cite{vgrnn}, we used a single-cell graphical gated recurrent unit (GGRU) for the function $\textrm{Update}$. 
The GGRU cell takes the concatenation of structural and time embeddings, $\textrm{CONCAT}(\mathbf{Z}_k, k_{\textrm{enc}})$, as input, and along with the previous states $\mathbf{S}_{k-1}$, it computes the updated states by applying message-passing over the current graph's structure $\mathbf{A}_k$. 
The time embeddings, $k_{\textrm{enc}}=\textrm{TimeEncoder}(k)$, are computed using the proposed time encoder in \cite{graphgps}. 
The GGRU allows the model to spread the state information across time-respecting paths.
The recurrent $\textrm{Update}$ applies to all the time steps $k=1, \dots, N$, with the initial state being the zero matrix $\mathbf{S}_0 = \mathbf{0}$. 
See Appendix \ref{subsec:GGRU} and \ref{subsec:TimeEncoder} for further details.
Algorithm \ref{alg:inference} shows the inference step of the model, which produces the state representation matrix $\mathbf{S}$, which can be used in link reconstruction or prediction settings.

\subsubsection{Decoder}
Provided the node states of the network $\mathbf{S}_{k}$ at time step $k$, the $\textrm{Decoder}$ function is responsible for reconstructing the network's structure, $\hat{\mathbf{A}}_k = P(\mathbf{A}_k | \mathbf{S}_{k})$:  
\begin{equation}\label{eq:linkrecon}
	\hat{\mathbf{A}}_k = \textrm{Decoder}(\mathbf{S}_{k}) \quad k=1, \dots, N.
\end{equation}
In our experiments, we implemented the $\textrm{Decoder}$ function by applying a linear layer to the state matrix, 
$\hat{\mathbf{Y}}_k = \textrm{Linear}_\textrm{dec}(\mathbf{S}_k)$, and computed the link reconstruction probabilities as the inner product of the resulting embeddings, i.e., $P(e_{ij}=1 | \mathbf{S}_k[i], \mathbf{S}_k[j]) = \sigma(\hat{\mathbf{Y}}_k[i]^{\top}\hat{\mathbf{Y}}_k[j])$ where $\sigma$ is the sigmoid function and $e_{ij}$ represents the edge between nodes $v_i$ and $v_j$.

\subsubsection{LinkPredictor}
For the dynamic link prediction task, the LinkPredictor function operates on the current representations of the temporal network to predict the structure of the temporal network at the next timestamp,
$\tilde{\mathbf{A}}_{k+1} = P(\mathbf{A}_{k+1} | \mathbf{S}_{k})$:
\begin{equation}\label{eq:linkpred}
	\tilde{\mathbf{A}}_{k+1} = \textrm{LinkPredictor}(\mathbf{S}_{k})\quad k=1, \dots, N - 1.
\end{equation}

Similar to the $\textrm{Decoder}$ function, we first linearly transform the current state matrix, $\tilde{\mathbf{Y}}_k = \textrm{Linear}_\textrm{pred}(\mathbf{S}_k)$, and use the inner product of the resulting embedding matrix $\tilde{\mathbf{Y}}_k$ as the link probabilities of the next time step: $P(e_{ij}=1 |\mathbf{S}_k[i], \mathbf{S}_k[j]) = \sigma(\tilde{\mathbf{Y}}_k[i]^{\top}\tilde{\mathbf{Y}}_k[j])$.

\subsubsection{PredictiveEncoder}\label{subsec:predictiveencoders}
We aim to learn node representations that capture features of the temporal network that span multiple time steps into the future. 
For this purpose, we employ the contrastive predictive coding (CPC) framework \cite{oord2018representation}, which maximizes the mutual information between the context sequence and future features in a contrastive way.
Intuitively, by leveraging the infoNCE learning objective of the CPC framework, node representations are motivated to encode information that is aligned with future characteristics of the temporal network. 
Thus, infoNCE can be viewed as a regularization strategy for the link prediction objective of the model.
Given the complexity of temporal network data, we define two types of CPC objectives to guide the learning process: local and global infoNCEs.
Local infoNCE maximizes the agreement between node representations and their future structural embeddings. 
Additionally, to regularize the graph-level representations over time, we define the global infoNCE objective, which encourages the graph-level representations to agree with their future graph-level structural embeddings.
This multi-scale treatment of network dynamics, considering both local and global views of the graph, leads to the learning of more informative representations.
Details on the calculation of the infoNCE loss can be found in Subsection \ref{subsec:cpc_section}.
\begin{algorithm}[t]
\caption{Inference}\label{alg:inference}
\begin{algorithmic}[1]
\Input Snapshot sequence $\{G_k=(\mathbf{X}_k, \mathbf{A}_k)\}_{k=1}^{N}$
\Output State matrix $\mathbf{S}$
\State $\mathbf{S}=0$
\For{$k=1$ to $N$}
\State Encode the current graph, $\mathbf{Z}_k = \textrm{Encoder}(\mathbf{X}_k, \mathbf{A}_k)$ 
\State Update the states, $\mathbf{S} = \textrm{Update}(\mathbf{Z}_k, \mathbf{A}_k, \mathbf{S}, k)$
\EndFor
\end{algorithmic}
\end{algorithm}

To implement the CPC objectives, we define the local and global predictive encoders that act on the node and graph-level representations and compute the predictions of future structural embeddings for each iteration of processing the snapshot sequence.
These feature predictions will be used in computing the CPC loss objective of the training.

For the local node-level CPC, at the time step $k$, we compute the local predictive encodings for future time steps $l=\{k+1, \dots, N\}$ using node representations $\mathbf{S}_k$ at the time step $k$, which is denoted as $\hat{\mathbf{Z}}^{(k)}_l$:
\begin{equation}
        \hat{\mathbf{Z}}^{(k)}_l = \textrm{LocalPredictiveEncoder}(\mathbf{S}_k, l_{\textrm{enc}}).
\end{equation}

Similarly, for the global graph-level CPC, at the time step $k$, we compute the global predictive encodings of future time steps $l=\{k+1, \dots, N\}$ using graph representations $\mathbf{s}_k$, which is denoted by $\hat{\mathbf{z}}^{(k)}_l$:
\begin{equation}
    \begin{aligned}    
        & \mathbf{s}_k = \textrm{ReadOut}(\mathbf{S}_k), \\
        & \hat{\mathbf{z}}^{(k)}_l  =
        \textrm{GlobalPredictiveEncoder}(\mathbf{s}_k, l_{\textrm{enc}}),
    \end{aligned}
\end{equation}
where $l_{\textrm{enc}}=\textrm{TimeEncoder}(l)$ is the encoded time step defined in \cite{graphmixer}, and the $\textrm{ReadOut}$ function is defined as $\textrm{ReadOut}(\mathbf{S}) = \frac{1}{n} \sum_{i \in \mathcal{V}} \mathbf{S}[i]$ which averages the node-level representation and returns the corresponding graph-level representations.
Note that for the $\textrm{ReadOut}$ function, any available graph pooling method can be employed to account for the dynamics of hierarchical representations within the CPC framework \cite{diffpool, sagpool, mewispool}.

In the experiments, the $\textrm{LocalPredictiveEncoder}$ function is implemented using a 2-layer MLP with a $\textrm{ReLU}$ activation function that acts on the concatenation of node representations and time step embedding, i.e., 
\[
\begin{aligned}
	\hat{\mathbf{Z}}^{(k)}_l=\textrm{MLP}_{\textrm{local}}(\textrm{CONCAT}(\mathbf{S}_k, l_\textrm{enc})). 
\end{aligned}
\]
Moreover, for the $\textrm{GlobalPredictiveEncoder}$, we apply a linear transformation to the concatenation of graph-level representations and time step embedding:
\[
\begin{aligned}
	\hat{\mathbf{z}}^{(k)}_l=\textrm{Linear}_{\textrm{global}}(\textrm{CONCAT}(\mathbf{s}_k,  l_\textrm{enc})).
\end{aligned}
\]
\begin{algorithm}[t]
\caption{Training forward pass}\label{alg:forward-pass}
\begin{algorithmic}[1] 
\Input Snapshot sequence $\{G_k=(\mathbf{X}_k, \mathbf{A}_k)\}_{k=1}^{N}$
\Output Node state matrix $\mathbf{S}$, set of structural embeddings $Z$, set of reconstructed adjacency matrices $\hat{A}$, set of predicted adjacency matrices $\tilde{A}$, set of local and global predictive encodings, $\hat{Z}, \hat{z}$

\State $\mathbf{S}=0$
\State Initialize $Z, \hat{A}, \tilde{A}, \hat{Z}, \hat{z}$ with empty lists
\For{$k=1$ to $N$}
\State Encode the current graph, $\mathbf{Z}_k = \textrm{Encoder}(\mathbf{X}_k, \mathbf{A}_k)$ 
\State Update the states, $\mathbf{S} = \textrm{Update}(\mathbf{Z}_k, \mathbf{A}_k, \mathbf{S}, k)$

\State Reconstruct the current graph, $\hat{\mathbf{A}}_k = \textrm{Decoder}(\mathbf{S}_{k})$ 
\State Predict the next graph, $\tilde{\mathbf{A}}_{k+1} = \textrm{LinkPredictor}(\mathbf{S}_{k})$
\State Get local predictive encoding for $l=\{k+1, \dots, N\}$:
\State $\hat{\mathbf{Z}}^{(k)}_l = \textrm{LocalPredictiveEncoder}(\mathbf{S}_k, l_{\textrm{enc}})$
\State Get global graph state, $\mathbf{s}_k = \textrm{ReadOut}(\mathbf{S}_{k})$
\State Get global predictive encoding for $l=\{k+1, \dots, N\}$:
\State $\hat{\mathbf{z}}^{(k)}_l  =
        \textrm{GlobalPredictiveEncoder}(\mathbf{s}_k, l_{\textrm{enc}})$
\State Append $\mathbf{Z}_k, \hat{\mathbf{A}}_k, \tilde{\mathbf{A}}_{k+1}$ to $Z, \hat{A}, \tilde{A}$
\State Append local and global encodings to $\hat{Z}, \hat{z}$
\EndFor
\end{algorithmic}
\end{algorithm}

\subsection{Learning}
The model parameters are trained by minimizing the multi-objective loss function using the stochastic gradient descent optimization algorithm. 
As mentioned in \ref{subsec:motivation}, the loss function encompasses three objectives, each of which corresponds to a desired aspect of learning node state representations of the temporal network:
\begin{equation}\label{eq:loss}
	\begin{aligned}
		L = L_{\textrm{pred}} + \alpha L_{\textrm{recon}} + \beta L_{\textrm{cpc}},
	\end{aligned}
\end{equation}
where $\alpha$ and $\beta$ are scalar hyperparameters used for weighting the relative importance of loss terms compared to the dynamic link prediction loss.

\subsubsection{Link prediction loss}
In this work, the main objective of learning node state representations is to extract useful information for predicting the structure of the temporal network at the next time step.
Hence, for the link prediction loss, we compute the binary cross-entropy (BCE) classification loss between the predicted and ground-truth graph structures for each prediction time step:
\begin{equation}\label{eq:linkpredloss}
	\begin{aligned}
		L_{\textrm{pred}} = \frac{1}{N-1}\sum_{k=1}^{N-1}\textrm{BCE}(\tilde{\mathbf{A}}_{k+1}, \mathbf{A}_{k+1}).
	\end{aligned}
\end{equation}

We take the average of all link prediction losses over indices $k=1, \dots, N - 1$ as the prediction loss of the model, $L_{\textrm{pred}}$. 

\subsubsection{Reconstruction loss}
Apart from the predictive power of the node states, our goal is to learn representations that capture information about the current structure of the temporal network. 
For this purpose, we add the reconstruction loss to the training objectives.
This objective is implemented as the graph autoencoder reconstruction loss \cite{gae} for all the snapshots in the input sequence, which is essentially the BCE loss between reconstructed and ground-truth adjacency matrices: 
\begin{equation}\label{eq:linkreconloss}
	\begin{aligned}
		L_{\textrm{recon}} = \frac{1}{N}\sum_{k=1}^N\textrm{BCE}(\hat{\mathbf{A}}_k, \mathbf{A}_k).
	\end{aligned}
\end{equation}

We also compute the average of all reconstruction losses as $L_{\textrm{recon}}$.

\subsubsection{Contrastive predictive coding loss} \label{subsec:cpc_section}
The final term in the model's training objective Eq.\ref{eq:loss} is the CPC loss, which was originally introduced in \cite{oord2018representation}. 
As described in Subsection \ref{subsec:predictiveencoders}, we employ local and global infoNCE losses to maximize the mutual information between node representations and future features of the temporal graph at local and global scales, respectively.   
These losses assist the model in achieving a balance between learning low-level features, which are beneficial for reconstruction and one-step prediction tasks, and extracting slow-varying features that characterize the temporal network over more extended periods.

The CPC loss of the model is computed as the average infoNCE loss over time steps $k=1, \dots, N-1$: 
\begin{equation}\label{eq:cpcloss}
	\begin{aligned}
		& L_\textrm{cpc}=\frac{1}{N-1}\sum_{k=1}^{N-1}\textrm{infoNCE}^{(k)} \quad k=1, \dots, N-1, \\
	\end{aligned}
\end{equation}
where for each predictive coding time step $k$, $\textrm{infoNCE}^{(k)}$ is computed as the sum of local and global noise contrastive losses from current time step $k$ to the final training time step $N$.
We have:
\begin{equation}
	\begin{aligned}
		& \textrm{infoNCE}^{(k)}=\sum_{l=k+1}^{N}\textrm{localNCE}^{(k)}_l + \textrm{globalNCE}^{(k)}_l. \\
	\end{aligned}
\end{equation}
\begin{figure}[t]
    \centering
    \includegraphics[width=1\columnwidth]{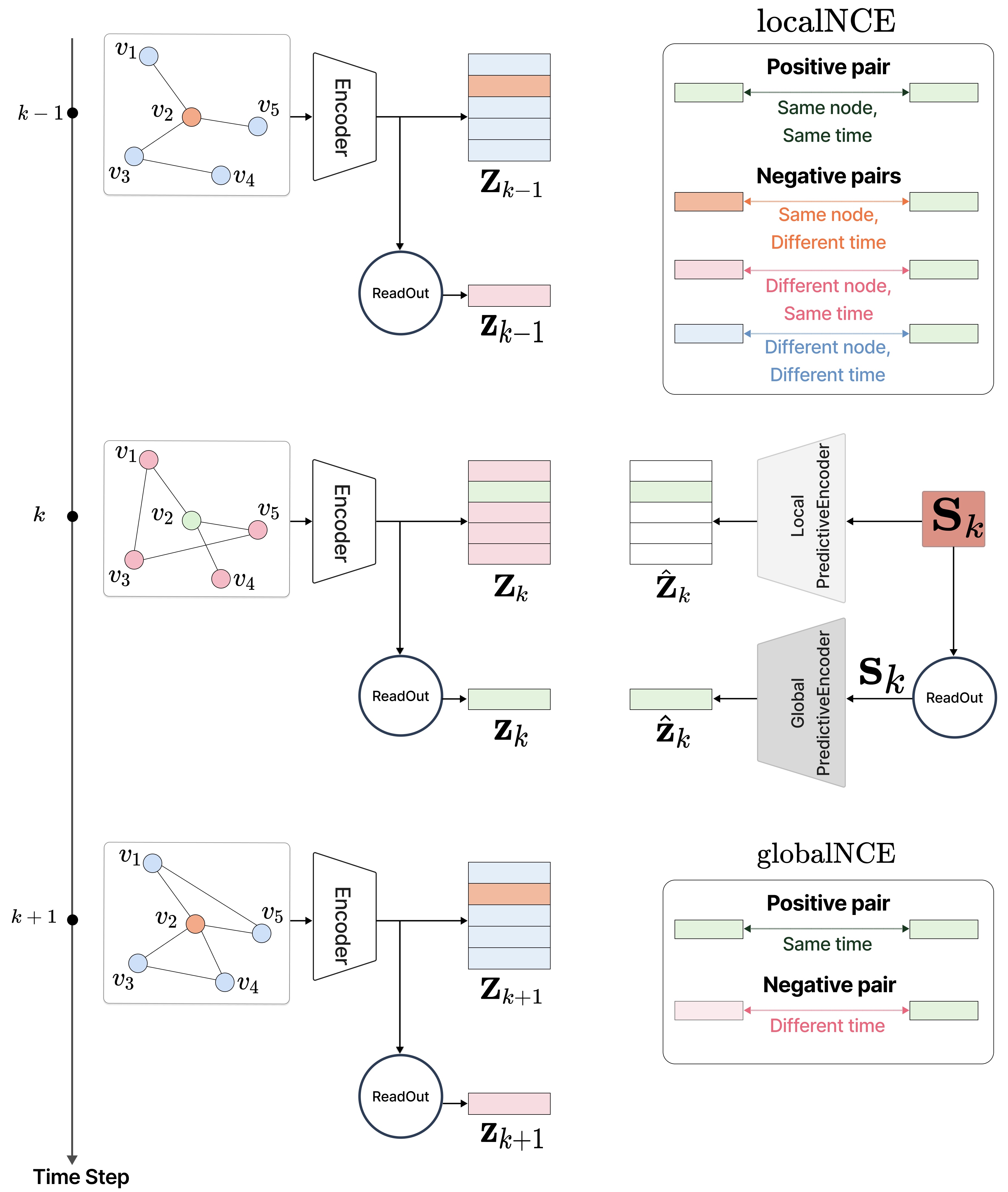} 
    \caption{Illustration of positive and negative sample pairs for local and global infoNCE losses. This example depicts positive and negative pairs for node the $v_2$ and graph $G_k$, corresponding to local and global losses. For localNCE, different negative samples defined in Eq.\ref{eq:negative_samples} are colored orange, pink, and blue; for globalNCE, negative samples are colored pink.}
    \label{fig:nagtive_samples}
\end{figure}

Local noise contrastive loss at time step $k$ with respect to the future time step $l$, denoted as $\textrm{localNCE}^{(k)}_l$, is computed at the node-level of the temporal network. 
By minimizing this term we maximize the mutual information between each node's representation, $\mathbf{S}_k[i]$, with the node's future features, $\mathbf{Z}_l[i]$ for $l=\{k+1, \dots, N\}$, which results in learning node representations that are more aligned with future characteristics of the temporal network \cite{oord2018representation}.
At the time step $k$, the loss term $\textrm{localNCE}^{(k)}_l$ for node $i$ with respect to time step $l$, is implemented as the binary cross-entropy loss between the positive pair $(\hat{\mathbf{Z}}^{(k)}_{l}[i], \mathbf{Z}_{l}[i])$ and the negative pairs  $(\hat{\mathbf{Z}}^{(k)}_{l}[i], \mathbf{Z}_{l^\prime}[i^{\prime}])$, where $(i^{\prime}, l^\prime)$ is drawn from the local negative sample distribution for node $i$ at time step $l$, i.e.,  $(i^{\prime}, l^{\prime})\sim P_\text{neg}(i, l)$.
Specifically, we have:
\begin{equation}
    \resizebox{0.5\textwidth}{!}{$
    \begin{aligned}
    & \text{localNCE}^{(k)}_l = \\
    & \frac{-1}{n} \sum_{i \in \mathcal{V}} \mathbb{E}_{P_\text{neg}(i, l)} \left(\log \frac{\exp\left(\hat{\mathbf{Z}}^{(k)}_{l}[i]^\top \mathbf{Z}_{l}[i]\right)}{\exp\left(\hat{\mathbf{Z}}^{(k)}_{l}[i]^\top \mathbf{Z}_{l}[i]\right) + \sum\limits_{(i^{\prime}, l^{\prime})\sim P_\text{neg}(i, l)}\exp\left(\hat{\mathbf{Z}}^{(k)}_{l}[i]^\top \mathbf{Z}_{l^\prime}[i^\prime]\right)}\right).
    \end{aligned}$}
\end{equation}
In a similar manner, the global noise contrastive loss at time step $k$ with respect to the time step $l$, denoted by $\textrm{globalNCE}^{(k)}_l$,  is computed at the global graph-level scale of the temporal network.
In particular, the loss term $\textrm{globalNCE}^{(k)}_l$ is computed using the graph-level representations of the temporal network, $\mathbf{s}_k$, and the graph-level structural encodings at the future time steps, $\mathbf{z}_{l}$ for $l=\{k+1, \dots, N\}$.
At the time step $k$ and with respect to time step $l$, the loss term $\textrm{globalNCE}^{(k)}_l$ is implemented as the binary cross-entropy loss between the positive pair $(\hat{\mathbf{z}}^{(k)}_{l}, \mathbf{z}_{l})$, and the negative pairs  $(\hat{\mathbf{z}}^{(k)}_{l}, \mathbf{z}_{l^\prime})$ where $l^\prime$ is drawn from the global negative sample distribution at time step $l$, i.e.,  $l^\prime\sim P_\text{neg}(l)$:
\begin{equation}
    \resizebox{0.45\textwidth}{!}{$
    \begin{aligned}
    & \text{globalNCE}^{(k)}_l = \\
    & -\mathbb{E}_{P_\text{neg}(l)}\left(\log \frac{\exp\left(\hat{\mathbf{z}}^{(k)}_{l}{}^\top \mathbf{z}_{l}\right)}{\exp\left(\hat{\mathbf{z}}^{(k)}_{l}{}^\top \mathbf{z}_{l}\right) + \sum\limits_{l^\prime\sim P_\text{neg}(l)}\exp\left(\hat{\mathbf{z}}^{(k)}_{l}{}^\top \mathbf{z}_{l^\prime}\right)}\right).
    \end{aligned}$}
\end{equation}

The local negative sample distribution, $P_\text{neg}(i, l)$, is implemented as the uniform probability distribution defined over the set of all negative samples, denoted as $\text{neg}_{(i, l)}$, that is associated to the node $i$ at time step $l$:
\[
\begin{aligned}
	\text{neg}_{(i, l)} = \{(i^\prime, l^\prime) \in \mathcal{V} \times \{1, \dots, N\}:
  (i^\prime \neq i) \; \textrm{OR} \; (l^\prime \neq l)\}.
\end{aligned}
\]

\begin{algorithm}[t]
\caption{Training step}\label{alg:training-step}
\begin{algorithmic}[1]
\Input Node state matrix $\mathbf{S}$, set of structural embeddings $Z$, set of reconstructed adjacency matrices $\hat{A}$, set of predicted adjacency matrices $\tilde{A}$, set of local and global predictive encodings, $\hat{Z}, \hat{z}$, loss function coefficients $\alpha, \beta$
\Output Model with updated parameters

\State Compute $L_\textrm{pred}$ Eq.\ref{eq:linkpredloss}
\State Compute $L_\textrm{recon}$ Eq.\ref{eq:linkreconloss}
\State Compute $L_\textrm{cpc}$ Eq.\ref{eq:cpcloss}
\State $L = L_{\textrm{pred}} + \alpha L_{\textrm{recon}} + \beta L_{\textrm{cpc}}$
\State Update model's parameters using backpropagation
\end{algorithmic}
\end{algorithm}
\begin{table}
    \caption{Dataset statistics.}
    \begin{center}
        \begin{tabular}{lcccc}
            \toprule
            {\textbf{Attribute}} 
            & \textbf{Enron} & \textbf{Colab} & \textbf{Facebook} \\
            \midrule
            \# Nodes & 184 & 315 & 663 \\
            
            \# Edges & 4,784 & 5,104 & 23,394 \\

            \# Time steps & 11 & 10 & 9
            \\
            \bottomrule
        \end{tabular}
    \end{center}
    \label{table:datasets}
\end{table}

The elements of negative samples set $\text{neg}_{(i, l)}$ consist of all other nodes in the temporal network across all time steps. More specifically, for node $i$ at time $l$, its set of negative samples can be partitioned into three subsets: 
\begin{equation}\label{eq:negative_samples}
    \begin{cases}
    \text{Same node, different time:} & (i=i^{\prime}, l \neq l^{\prime}),  \\
    \text{Different node, same time:} &(i \neq i^{\prime}, l=l^{\prime}), \\
    \text{Different node, different time:} & (i \neq i^{\prime}, l \neq l^{\prime}). 
    \end{cases}
\end{equation}

Similarly, the global negative sample distribution, $P_\text{neg}(l)$, is defined as the uniform probability distribution over the set of all global negative samples corresponding to time step $l$:
\begin{equation}
	\text{neg}_{(l)} = \{l^\prime \in \{1, \dots, N\}: l^\prime \neq l\},
\end{equation}
i.e., the set of all time steps other than the time step $l$.
These negative samples are illustrated in Figure \ref{fig:nagtive_samples}.

\section{Experiments}\label{sec:experiments}
In this section, we present a comprehensive set of experiments demonstrating the effectiveness of the proposed teneNCE model. 
All the implementations are developed in Python, using the Pytorch \cite{paszke2017automatic} and Pytorch Geometric \cite{fey2019fast} libraries.
For all the datasets, we have used the ADAM optimizer with the learning rate of $10^{-3}$ and the weight decay coefficient of $5\times10^{-4}$.
We further configured the ReduceLROnPlateau learning rate scheduler in training with a factor of $0.8$.
We also found that an embedding dimension of $256$ performs best for all datasets. A grid search was performed to find the optimal values for $\alpha$ and $\beta$ in Eq \ref{eq:loss}. The best values for $\alpha$ are 1, 2, and 4 for the Enron, COLAB, and Facebook datasets, respectively, while the best values for $\beta$ are 1, 4, and 2.

\begin{table*}[t] 
\fontsize{7.5pt}{9pt}\selectfont
    \caption{The empirical results for the random selection of positive and negative edges. The results represent the mean and standard deviation of evaluation metrics, which are obtained from five separate training runs of each model.}
    \begin{center}

        \begin{tabularx}{\textwidth}{lXXXXXXXXX}
        \toprule
        \multirow{2}{*}{\textbf{Model}} & \multicolumn{3}{c}{\textbf{Enron}} & \multicolumn{3}{c}{\textbf{COLAB}} & \multicolumn{3}{c}{\textbf{Facebook}} \\
        \cmidrule(lr){2-4} \cmidrule(lr){5-7} \cmidrule(lr){8-10}
        & \multicolumn{1}{c}{\textbf{AUC}} & \multicolumn{1}{c}{\textbf{AP}} & \multicolumn{1}{c}{\textbf{MRR}} & \multicolumn{1}{c}{\textbf{AUC}} & \multicolumn{1}{c}{\textbf{AP}} & \multicolumn{1}{c}{\textbf{MRR}} & \multicolumn{1}{c}{\textbf{AUC}} & \multicolumn{1}{c}{\textbf{AP}} & \multicolumn{1}{c}{\textbf{MRR}} \\
        
        \midrule
            VGRNN & 92.68 $\pm$ 0.5 & 93.04 $\pm$ 0.3 & 29.02 $\pm$ 0.3 & 
            86.56 $\pm$ 0.4 & 88.52 $\pm$ 0.5 & 27.20 $\pm$ 1.5 
            & 89.44 $\pm$ 0.1 & 88.82 $\pm$ 0.3 & 12.06 $\pm$ 0.2\\
            
            DySAT & 90.64 $\pm$ 0.9 & 90.33 $\pm$ 1.2 & 19.00 $\pm$ 0.6
            & 86.10 $\pm$ 0.3 & 89.32 $\pm$ 0.3 & 26.19 $\pm$ 0.6
            & 89.95 $\pm$ 0.3 & 89.56 $\pm$ 0.3 & 12.74 $\pm$ 0.2
            \\
            EvolveGCN-H & 87.99 $\pm$ 0.6 & 87.74 $\pm$ 1.3 & 23.23 $\pm$ 1.7
            & 81.48 $\pm$ 1.4 & 82.95 $\pm$ 1.3 & 17.11 $\pm$ 1.1
            & 82.53 $\pm$ 1.2 & 79.89 $\pm$ 1.7 & 5.68 $\pm$ 1.0
            \\
            EvolveGCN-O & 88.87 $\pm$ 0.5 & 89.13 $\pm$ 0.6 & 25.24 $\pm$ 1.2 
            & 80.63 $\pm$ 0.9 & 82.79 $\pm$ 0.9 & 18.17 $\pm$ 2.6
            & 82.85 $\pm$ 0.4 & 81.39 $\pm$ 0.6 & 6.73 $\pm$ 0.6
            \\
            Euler & 91.28 $\pm$ 1.6 & 91.47 $\pm$ 1.6 & 27.45 $\pm$ 2.6
            & 85.93 $\pm$ 1.0 & 87.51 $\pm$ 0.8 & 20.91 $\pm$ 2.2
            & 87.89 $\pm$ 1.1 & 85.84 $\pm$ 1.5 & 8.49 $\pm$ 0.8
            \\
            \midrule
            
            teneNCE & \textbf{93.54 $\pm$ 0.4} & \textbf{93.65 $\pm$ 0.2 } & \textbf{31.5 $\pm$ 0.6 }
            
            & \textbf{88.25 $\pm$ 0.7} & \textbf{90.45 $\pm$ 0.6} & \textbf{33.97 $\pm$ 0.6}
            
            & \textbf{91.03 $\pm$ 0.1} & \textbf{90.32 $\pm$ 0.2} & \textbf{16.23 $\pm$ 0.2}
            \\
            \bottomrule
        \end{tabularx}
    \end{center}
    \label{table:results}
\end{table*}
\begin{figure*}[t] 
    \centering
    \includegraphics[width=1.8\columnwidth]{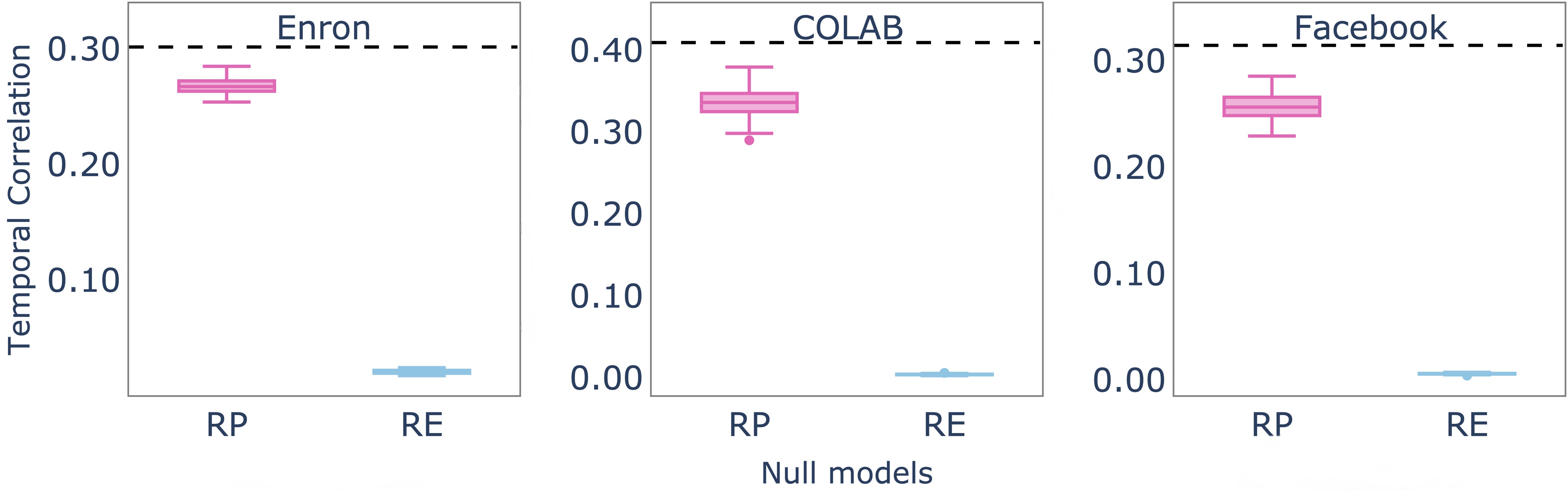} 
    \caption{Visualization of the comparison between the temporal correlation coefficient values for the datasets in \ref{subsec:datasets} and their corresponding randomized versions. We have used two randomization models, namely Randomly Permuted Times (RP) and Randomized Edges (RE). For each null model, we have sampled 100 randomized versions. The box plots summarize the distribution of temporal correlation coefficients of the randomized samples. The figure demonstrates that the null models have a lower temporal correlation than the original data. This suggests that capturing temporal information of dynamic graphs is essential for effectively modeling such data.}
    \label{fig:tcc}
\end{figure*}
\subsection{Datasets}\label{subsec:datasets}
We use the following datasets to evaluate the performance of the proposed model. 
The summary statistics for these datasets are presented in Table \ref{table:datasets}.
\begin{table*}[t]
\fontsize{9pt}{9pt}\selectfont
    \caption{The empirical results for evaluation subsets Rand-Pos/Hist-Neg, Hist-Pos/Rand-Neg, and Hist-pos/Hist-Neg, as defined in \ref{subsec:task-eval}.}
    \begin{center}
        \begin{tabular}{l|lccccccc}
            \toprule
            \multirow{3}{*}{\textbf{Dataset}} & 
            \multirow{3}{*}{\textbf{Model}} & 
            \multicolumn{2}{c}{\textbf{Rand-Pos/Hist-Neg}} & 
            \multicolumn{2}{c}{\textbf{Hist-Pos/Rand-Neg}} & 
            \multicolumn{2}{c}{\textbf{Hist-Pos/Hist-Neg}}  \\

\\[-0.5em]
			\cline{3-8}
\\[-0.5em]		    
            & & \textbf{AUC} & \textbf{AP} 
            & \textbf{AUC} & \textbf{AP} 
            & \textbf{AUC} & \textbf{AP} 
            \\
            
            \midrule
             \multirow{6}{*}{Enron} & VGRNN
            & 64.03 $\pm$ 0.1 & 66.92 $\pm$ 0.3
            & 96.30 $\pm$ 0.2 & 95.81 $\pm$ 0.1
            & 72.32 $\pm$ 0.3 & 73.28 $\pm$ 0.6
 \\
            
            & DySAT 
            & 58.30 $\pm$ 1.4 & 57.54 $\pm$ 1.0
            & 93.73 $\pm$ 0.7 & 92.88 $\pm$ 0.9
            & 63.43 $\pm$ 0.4 & 61.00 $\pm$ 0.7
\\
            
            & EvolveGCN-H 
            & 63.77 $\pm$ 0.4 & 64.04 $\pm$ 0.6
            & 91.96 $\pm$ 0.8 & 90.95 $\pm$ 1.5
            & 70.41 $\pm$ 0.2 & 68.86 $\pm$ 0.2
\\
            
            & EvolveGCN-O
            & 63.82 $\pm$ 0.4 & 64.65 $\pm$ 0.7
            & 93.56 $\pm$ 0.7 & 92.61 $\pm$ 0.5
            & 71.33 $\pm$ 1.3 & 70.30 $\pm$ 1.5
\\
            
            & Euler
            & 62.13 $\pm$ 1.4 & 64.20 $\pm$ 2.0
            & 95.08 $\pm$ 1.3 & 94.57 $\pm$ 1.3
            & 70.59 $\pm$ 1.7 & 70.50 $\pm$ 2.7
\\
            
            & teneNCE 
            & \textbf{65.23 $\pm$ 0.2} & \textbf{68.32 $\pm$ 0.7}
            & \textbf{96.81 $\pm$ 0.1} & \textbf{96.20 $\pm$ 0.2}
            & \textbf{73.95 $\pm$ 0.5} & \textbf{74.29 $\pm$ 0.8}
\\
            
            \midrule
            \multirow{6}{*}{COLAB} & VGRNN
            & 52.44 $\pm$ 0.8 & 55.38 $\pm$ 1.2 
            & 96.52 $\pm$ 0.5 & 96.08 $\pm$ 0.8 
            & 69.04 $\pm$ 1.2 & 66.25 $\pm$ 1.8 
\\
            
            & DySAT 
            & 49.77 $\pm$ 0.3 & 55.51 $\pm$ 0.6
            & 97.38 $\pm$ 0.4 & 97.56 $\pm$ 0.3
            & 68.86 $\pm$ 0.3 & 67.73 $\pm$ 1.3
\\
            
            & EvolveGCN-H 
            & 57.07 $\pm$ 1.1 & 57.72 $\pm$ 1.3
            & 91.11 $\pm$ 1.0 & 90.63 $\pm$ 1.0
            & 69.48 $\pm$ 0.4 & 66.08 $\pm$ 0.3
\\
            
            & EvolveGCN-O
            & \textbf{57.55 $\pm$ 0.6} & \textbf{59.00 $\pm$ 0.6}
            & 90.20 $\pm$ 0.7 & 90.42 $\pm$ 0.8
            & 70.70 $\pm$ 1.0 & 68.76 $\pm$ 1.6
\\
            
            & Euler
            & 51.91 $\pm$ 1.0 & 54.54 $\pm$ 0.9
            & 95.07 $\pm$ 0.5 & 94.54 $\pm$ 0.7
            & 65.75 $\pm$ 1.4 & 63.47 $\pm$ 1.8
\\
            
            & teneNCE 
            & 55.29 $\pm$ 0.7 & 58.79 $\pm$ 1.2
            & \textbf{97.88 $\pm$ 0.3} & \textbf{97.69 $\pm$ 0.3}
            & \textbf{72.96 $\pm$ 1.1} & \textbf{69.95 $\pm$ 1.6}
\\
            
            \midrule
            \multirow{6}{*}{Facebook} & VGRNN
            & 52.08 $\pm$ 0.2 & 53.57 $\pm$ 0.1 
            & 93.19 $\pm$ 0.2 & 92.15 $\pm$ 0.3 
            & 59.63 $\pm$ 0.2 & 59.12 $\pm$ 0.3 
\\
            
            & DySAT 
            & 52.06 $\pm$ 0.2 & 53.53 $\pm$ 0.3
            & 93.58 $\pm$ 0.2 & 92.84 $\pm$ 0.2
            & 59.76 $\pm$ 0.4 & 59.19 $\pm$ 0.5
\\
            
            & EvolveGCN-H 
            & 53.52 $\pm$ 0.7 & 52.86 $\pm$ 0.6
            & 85.34 $\pm$ 1.4 & 82.68 $\pm$ 1.8
            & 58.38 $\pm$ 0.7 & 56.65 $\pm$ 0.5
\\
            
            & EvolveGCN-O
            & 53.32 $\pm$ 0.3 & 52.84 $\pm$ 0.4
            & 86.02 $\pm$ 0.5 & 84.46 $\pm$ 0.9
            & 58.64 $\pm$ 0.6 & 56.85 $\pm$ 0.6
\\
            
            & Euler
            & 50.65 $\pm$ 0.5 & 52.19 $\pm$ 0.4
            & 90.45 $\pm$ 1.2 & 88.65 $\pm$ 1.6
            & 56.58 $\pm$ 0.6 & 56.81 $\pm$ 0.5
\\
            
            & teneNCE 
            & \textbf{54.65 $\pm$ 0.3} & \textbf{55.75 $\pm$ 0.5}
            & \textbf{94.21 $\pm$ 0.1} & \textbf{93.27 $\pm$ 0.2}
            & \textbf{62.39 $\pm$ 0.5} & \textbf{61.64 $\pm$ 0.6}
\\
            
            \bottomrule
        \end{tabular}

    \end{center}
    \label{table:enron_colab_facebook_results} 
    \end{table*}

\textbf{Enron.}
The Enron dataset captures the email communication network of the Enron Corporation, providing a historical perspective on interactions among employees. 
In this dataset, nodes represent employees, and edges correspond to email exchanges between colleagues \cite{priebe2005scan}. 

\textbf{Collaboration (COLAB).}
This dataset consists of collaboration data among 315 authors, where each author is represented as a node in the dynamic graph and edges correspond to co-authorship relationships \cite{rahman2016link}.

\textbf{Facebook.}
The Facebook dataset represents the social connections among users on the Facebook platform. 
It contains a dynamic graph of friendships and interactions, facilitating research on social network dynamics, information diffusion, and community structure within an online social network \cite{viswanath2009evolution}.

We evaluate our proposed method, teneNCE and the baselines on the last three snapshots of each dataset.
We computed the temporal correlation coefficients of the corresponding temporal network to provide insight into each dataset's temporal evolution.
As defined in \cite{toolbox}, the temporal correlation coefficient is computed as
\begin{equation}
	C=\frac{1}{n} \sum_i C_i,
\end{equation}
where, 
\begin{equation}
C_i=\frac{1}{N-1} \sum_{k=1}^{N-1} \frac{\sum_j \mathbf{A}_k[i, j] \mathbf{A}_{k+1}[i, j]}{\sqrt{\left[\sum_j \mathbf{A}_k[i, j]\right]\left[\sum_j \mathbf{A}_{k+1}[i, j]\right]}},
\end{equation}
is the temporal correlation coefficient of each node in the temporal network. 
The dashed lines represented in the box plots of Figure \ref{fig:tcc} indicate the temporal correlation coefficient of each dataset.

Following \cite{toolbox}, we compared the temporal correlation coefficient of each dataset with its respective randomized versions.
For the randomization process, we employed two null models: Randomized Edges (RE) and Randomly Permuted Times (RP).
The randomized models RE and RP destroy the structural and temporal information of the snapshot sequence, respectively \cite{gauvin1806randomized, toolbox}.
The difference between the temporal correlation coefficient of the data and their corresponding randomized versions demonstrates the richness of temporal information present within these networks.
This supports the application of temporal models to encode these networks, as opposed to relying on static models.

\begin{figure}[t]
\centering
 \includegraphics[width=0.76\columnwidth]{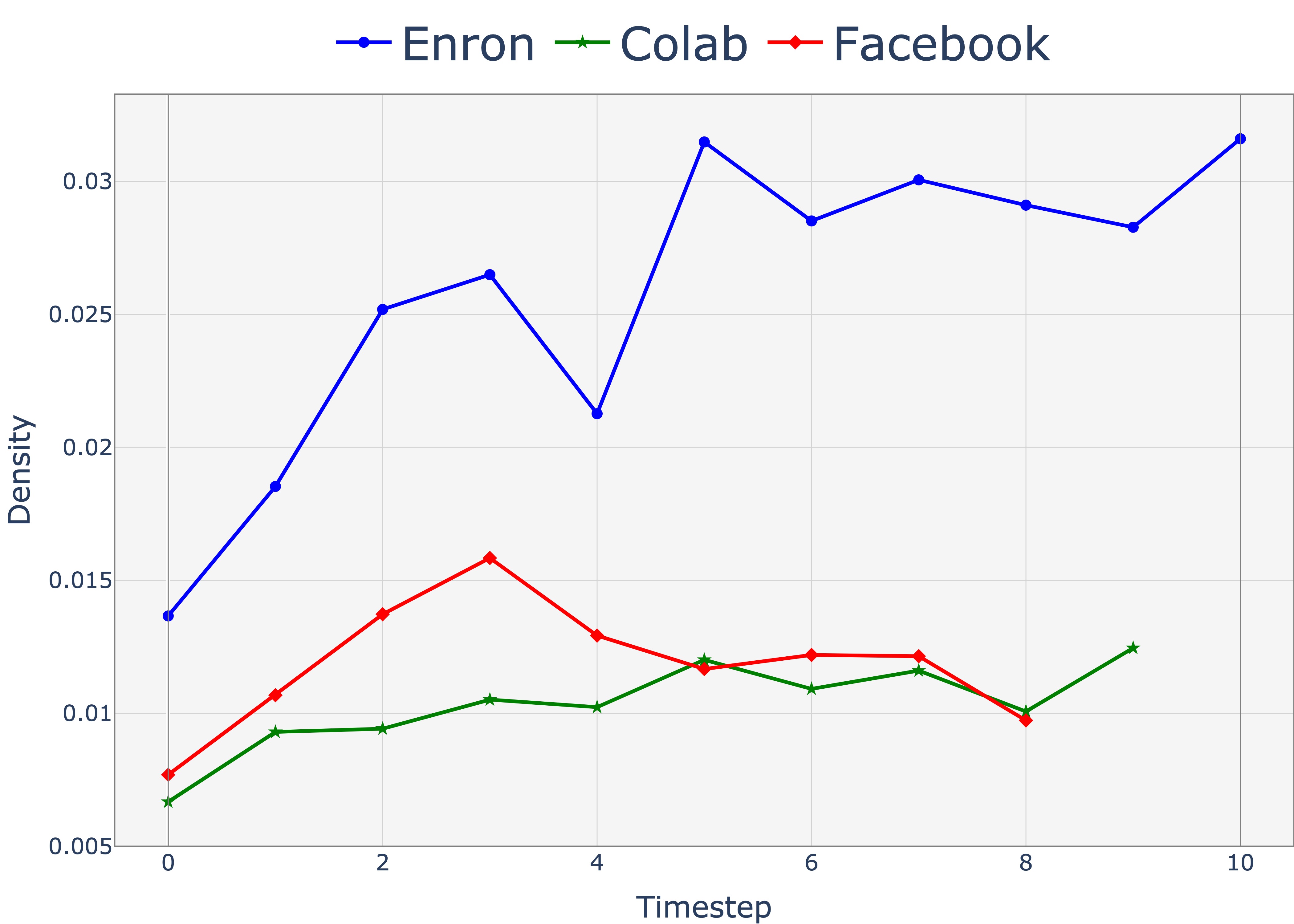}
\caption{The density values of snapshot sequences over time. The figure shows the variation of each graph's structure over time.}
\label{fig:density}
\end{figure}

Additionally, Figure \ref{fig:density} illustrates the evolution of graph density over time in the Enron, COLAB, and Facebook datasets.
Graph density is defined as the ratio of the number of existing edges in a graph to the maximum possible number of edges. 
This ratio provides a measure of the density of the graph in terms of edge connectivity.
As depicted in Figure \ref{fig:density}, the overall connectivities of the datasets vary over time, highlighting the necessity to consider temporal information in modeling. 

\subsection{Baselines}
We compared our method with existing GNN-based models over discrete-time dynamic graphs, including VGRNN \cite{vgrnn}, DySAT \cite{dysat}, EvolveGCN \cite{evolvegcn}, and Euler \cite{Euler}. 
A detailed introduction of these baselines can be found in Section \ref{sec:related_work}. 

\subsection{Tasks and Evaluation}\label{subsec:task-eval}
In this study, our focus is on the graph machine learning task of dynamic link prediction.
Specifically, we are interested in estimating the conditional probability of observing an edge between the nodes $v_i$ and $v_j$ at the time step $k$, given the historical data of the dynamic graph, i.e., $P(\mathbf{A}_{k+1}[i, j] = 1| G_1, \dots, G_k)$. 

As it is discussed in Section \ref{sec:method}, we employ the teneNCE model to encode the historical interactions of each node $v_i$ into a state vector $\mathbf{S}_k[i]$ that represents the temporal-topological information of the node. 
We then transform the states into an embedding space where conditional probabilities of edges are defined as the sigmoid function acting on the inner products of embeddings, i.e.,  $P(\mathbf{A}_{k+1}[i, j] = 1| G_1, \dots, G_k) = \sigma(\tilde{\mathbf{Y}}_k[i]^\top\tilde{\mathbf{Y}}_k[j])$, where $\tilde{\mathbf{Y}}_k=\textrm{Linear}(\mathbf{S}_k)$.
\begin{figure}[t]
    \centering
    \includegraphics[width=1\columnwidth]{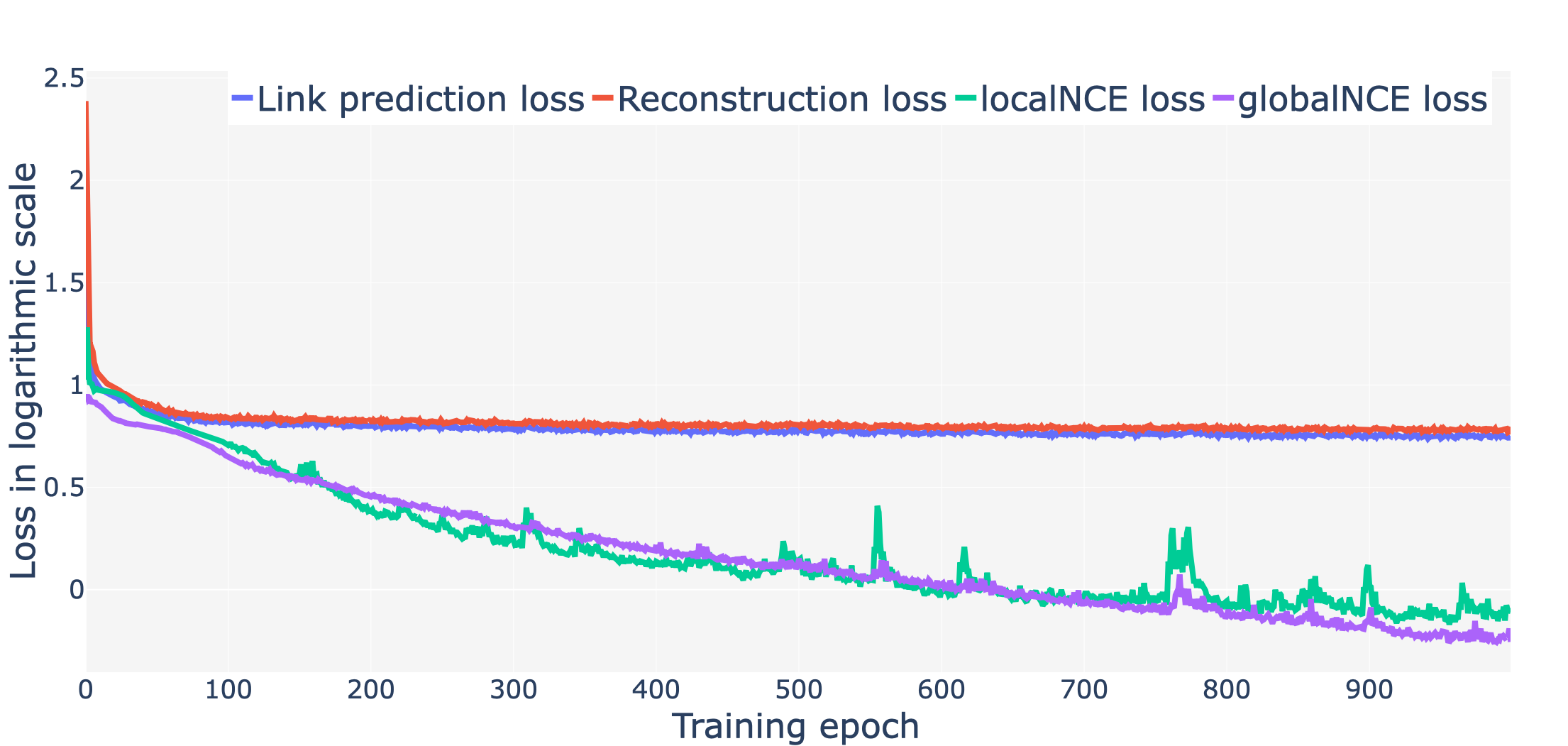} 
    \caption{Visualization of teneNCE loss minimization for the Enron dataset. The figure shows the evolution of loss values on a logarithmic scale throughout the training process.}
    \label{fig:loss-curves}
\end{figure}
Our main goal is to predict the positive edges in the graph in the future time steps. For this prediction, we consider a set of negative edges to train the model based on Eq. \ref{eq:loss} Positive edges are typically selected randomly from future time steps, but we define four strategies to assess the memorization and generalization capabilities of different models, following the evaluation procedure proposed in \cite{poursafaei2022towards}.
\begin{table*}[t]
\fontsize{9pt}{10pt}\selectfont
    \caption{Ablation study of teneNCE objective. Starting with the baseline link prediction loss, we sequentially incorporate the reconstruction, localNCE, and globalNCE losses at each stage. The figures indicate the contribution of each loss term on the model’s performance in dynamic link prediction.}
    \begin{center}
        \begin{tabular}{lcccccc}

            \toprule
            \multirow{2}{*}{\textbf{Loss}} & \multicolumn{2}{c}{\textbf{Enron}} & \multicolumn{2}{c}{\textbf{COLAB}} & \multicolumn{2}{c}{\textbf{Facebook}} \\
            \cmidrule(lr){2-3} \cmidrule(lr){4-5} \cmidrule(lr){6-7}
            & \textbf{AUC} & \textbf{AP} 
            & \textbf{AUC} & \textbf{AP} 
            & \textbf{AUC} & \textbf{AP} 
            \\
            \midrule
           Link Prediction
            & 91.28 $\pm$ 0.4 & 92.04 $\pm$ 0.2
            & 85.58 $\pm$ 0.5 & 88.29 $\pm$ 0.7
            & 87.51 $\pm$ 0.2 & 87.42 $\pm$ 0.2
            \\
           Link Prediction + Reconstruction
            & 92.03 $\pm$ 0.3 & 92.63 $\pm$ 0.5
            & 85.74 $\pm$ 0.7 & 88.44 $\pm$ 0.5
            & 87.69 $\pm$ 0.3 & 88.02 $\pm$ 0.2
            \\
           Link Prediction + Reconstruction + localNCE
            & 92.75 $\pm$ 0.5 & 92.85 $\pm$ 0.5
            & 87.81 $\pm$ 0.2 & 89.48 $\pm$ 0.3
            & 90.36 $\pm$ 0.1 & 89.67 $\pm$ 0.3
            \\
            \textbf{teneNCE loss (Eq.\ref{eq:loss})}
            & \textbf{93.54 $\pm$ 0.4} & \textbf{93.65 $\pm$ 0.2}  & 
            \textbf{88.25 $\pm$ 0.7} & \textbf{90.45 $\pm$ 0.6} &
            \textbf{91.03 $\pm$ 0.1} & \textbf{90.32 $\pm$ 0.2}
            \\
            \bottomrule
        \end{tabular}
    \end{center}
    \label{table:ablation_results}
\end{table*}
For existing positive edges at time step $k+1$, we define two subsets: \textit{random positive}, and \textit{historical positive} edges.
As the name implies, random positive edges are sampled at random from the edge set at the prediction time step, denoted as ${E}_{k+1}$.
However, historical positive edges refer to links that are observed in both the historical data and at the prediction time step $k+1$:
\begin{equation}
    \{(v_i, v_j):  (v_i, v_j) \in {E}_{k+1}, \; (v_i, v_j) \in \cup_{l=1}^k {E}_{l} \}.
\end{equation}

For the prediction of the absence of an edge at time step $k+1$, we further classified negative links into two subsets, namely, \textit{random negative} and \textit{historical negative} edges.
Similar to positive samples, random negative edges are randomly sampled from links that are inactive at the prediction time step. 
Furthermore, historical negatives refer to those links that were active in the historical data but have been removed at the prediction time step:
\begin{equation}
    \{(v_i, v_j):  (v_i, v_j) \notin {E}_{k+1}, \; (v_i, v_j) \in \cup_{l=1}^k {E}_{l}\}.
\end{equation}

The defined positive and negative edges form four evaluation subsets that we abbreviate as follows: Rand-Pos/Rand-Neg, Rand-Pos/Hist-Neg, Hist-Pos/Rand-Neg, and Hist-Pos/Hist-Neg.
The results for the Rand-Pos/Rand-Neg subset are presented in Table \ref{table:results}, providing a general comparison across various models.
For a more detailed comparison, the results for the other three test setups are presented in Table \ref{table:enron_colab_facebook_results}.
Our intuition is that a model with ample memorization capacity can more accurately predict the Hist-Pos/Rand-Neg links. 
On the other hand, a model with relatively more generalization power will predict the Rand-Pos/Hist-Neg edges with greater accuracy.

To measure the performance of the teneNCE model in comparison to baseline methods, we compute standard evaluation metrics commonly used in the dynamic link prediction tasks. 
These metrics include Average Precision (AP), AUC score (Area Under the ROC Curve), and Mean Reciprocal Rank (MRR). Detailed descriptions of these evaluation metrics are provided in the Appendix \ref{subsec:eval_metrics}.

\subsection{Performance Comparison}
For an overall assessment of the performance of various models on each dataset, we initially computed the evaluation metrics for the Rand-Pos/Rand-Neg test subset.
The results are provided in Table \ref{table:results} demonstrating enhancements over current methods across all datasets for all evaluation metrics.
Note that all experiments are conducted five times, and the mean and standard deviation of results are reported.
We observe that the teneNCE model achieves gains of up to 6\% MRR in the COLAB dataset when compared to the second-best baseline. 
VGRNN is typically the second-best model, which shows the power of VAEs in temporal modeling.
We also conducted paired t-test hypothesis testing to confirm the significance of numerical improvements \cite{weiss2005introductory}.
The calculated p-values suggest that the results hold statistical significance. For example, in the Enron dataset, the p-values for comparing the AUC and AP scores of teneNCE and VGRNN are $10^{-4}$ and $1.5 \times 10^{-3}$, respectively, leading to the rejection of the null hypothesis in the paired t-test.

Additionally, we assessed the performance of all models on the remaining three specific test setups as defined in \ref{subsec:task-eval}.
The results are presented in Table \ref{table:enron_colab_facebook_results}.
Apart from the COLAB dataset, where teneNCE’s performance is the second best after EvolveGCN-O, teneNCE has outperformed other models in different configurations and datasets.
The findings highlight the effectiveness of teneNCE in modeling the historical data and generalizing to unseen new links, relative to other methods. 

\subsection{Ablation study}
In this subsection, we aim to assess the contribution of the additional reconstruction and infoNCE loss terms to the teneNCE model’s performance.
To achieve this, we initially train the model using solely the baseline dynamic link prediction loss.
Subsequently, we iteratively augment the loss objective with additional reconstruction, local, and global infoNCE loss functions. 
At each stage, we compare the performance of the trained model against the baseline model and previous configurations.

The results are presented in Table \ref{table:ablation_results}.
We can observe the impact of each loss term on the baseline and previous measurements.
This study validates our motivations in formulating the training objective of teneNCE.
We observe a monotonic increase in the model’s performance when we enhance the baseline link prediction loss with additional loss terms.
In summary, we notice an average increase of $2.8$ in AUCs and $2.2$ in APs across the three datasets.

Specifically, we initially quantify the benefit of jointly optimizing prediction and reconstruction loss functions.
The reconstruction loss function ensures that the model captures the structural information necessary for reconstructing the graph from node states.
On average, adding the reconstruction objective to the baseline link prediction loss has a minor positive impact across datasets. 
However, the reconstruction loss aids the model in extracting representations that are beneficial in numerous applications, such as data imputation for connecting missing edges or densifying the input graph for improved information flow in subsequent machine-learning tasks. 

The primary enhancement in the baseline model’s performance can be attributed to the incorporation of both local and global infoNCE losses.
This indicates that the features obtained via the contrastive predictive coding technique are also advantageous for prediction in the data domain.

Figure \ref{fig:loss-curves} illustrates the progression of loss values for the teneNCE model throughout the training optimization process on the Enron dataset.
During the initial iterations of training, the link prediction and reconstruction losses are minimized. 
However, the infoNCE losses continue to decrease throughout the optimization process, underscoring the efficacy of self-supervised objectives in learning representations for temporal networks.

\section{Conclusion}\label{sec:conc}
In this study, we tackled the representation learning problem for temporal networks.
While temporal networks are typically characterized as contact sequences over continuous time, in this work, we modeled the discretized versions of these networks, also known as snapshot sequences.
This strategy balances the trade-off between computational complexity and precise modeling.

We introduced an architecture that models information propagation in temporal networks along time-respecting paths.
Our proposed architecture produces node representations that encode current and historical snapshots of the temporal network while maximizing the mutual information between the learned representations and the future network features.
This is accomplished by minimizing a multi-objective training loss function that combines prediction, reconstruction, and infoNCE loss terms.

We demonstrated that our proposed method outperforms existing models on discrete-time dynamic graphs, using the Enron, COLAB, and Facebook datasets as benchmarks.
This suggests that considering the complexity of temporal networks in real-world scenarios, self-supervised training with multiple objectives can assist models in deriving enhanced representations that capture multiple aspects of the underlying data.

For future research, we plan to apply the teneNCE architecture to supervised tasks over temporal networks, such as recommendation systems. 
Furthermore, we aim to use the proposed method in problems related to optimizing dynamical systems over dynamic graphs. 
This includes tasks such as planning and determining the optimal control over evolving networks, with practical applications such as those found in disease transmission networks.

{
\small
\bibliographystyle{ieee_fullname}
\bibliography{ref}
}
\newpage
\section{Appendix}\label{sec:appendix}
\subsection{Notations}\label{subsec:notations}
Table \ref{table:notations} provides a comprehensive list of the symbols and notations used throughout this paper. Each entry includes a brief description to clarify its meaning and context within the study. 

\subsection{Graphical Graph Recurrent Unit}\label{subsec:GGRU}
The graphical gated recurrent unit (GGRU), which is used to implement the function \textrm{Update} in Eq.\ref{eq:update}, is defined as:

\begin{equation}
	\begin{aligned}
		& \mathbf{R}_k = \sigma(\textrm{GConv}_{\text{reset,z}}(\mathbf{Z}_k, \mathbf{A}_k) + \textrm{GConv}_{\text{reset,s}}(\mathbf{S}_k, \mathbf{A}_k)), \\
		& \mathbf{U}_k = \sigma(\textrm{GConv}_{\text{update,z}}(\mathbf{Z}_k, \mathbf{A}_k) + \textrm{GConv}_{\text{update,s}}(\mathbf{S}_k, \mathbf{A}_k)), \\
		& \mathbf{\tilde{S}}_k = \tanh(\text{\footnotesize \textrm{GConv}}_{\text{cand,z}}(\mathbf{Z}_k, \mathbf{A}_k) + \mathbf{R}_k \odot \text{\footnotesize \textrm{GConv}}_{\text{cand,s}}(\mathbf{S}_k, \mathbf{A}_k)),\\
		& \mathbf{S}_{k+1} = (\mathbf{1}-\mathbf{U}_k) \odot \mathbf{\tilde{S}}_k + \mathbf{U}_k \odot \mathbf{S}_k.						
	\end{aligned}
\end{equation}

In the above equation, $\textrm{GConv}(.)$ denotes the graph convolutional layer introduced in \cite{gcn}, and $\odot$ denotes the element-wise multiplication.
$\mathbf{A}_k \in \mathbb{R}^{n \times n}$ is the adjacency matrix of the temporal network at time step $k$.
$\mathbf{Z}_k \in \mathbb{R}^{n \times d_{\textrm{enc}}}$ is the graph's embedding matrix, and $\mathbf{S}_k \in \mathbb{R}^{n \times d_{\textrm{state}}}$ is the node state matrix at time step $k$.
$\mathbf{R}_k, \mathbf{U}_k, \mathbf{\tilde{S}}_k, \mathbf{S}_{k+1} \in \mathbb{R}^{n \times d_{\textrm{state}}}$, are reset gates, update gates, candidate states and updated states, respectively. 

\subsection{Time Embedding Function}\label{subsec:TimeEncoder}
We used the time embedding function introduced in \cite{graphmixer}.
The function is defined as $\textrm{TimeEncoder(t)} = \cos(t \mathbf{\omega})$, which utilizes features $\mathbf{\omega}=\{\alpha^{\frac{−(i -1)}{\beta}} \}_{i=1}^{d_{\textrm{time}}}$ to encode each timestamps into a $d$-dimensional vector.
More specifically, the authors first map each $t$ to a vector with monotonically
exponentially decreasing values $t\mathbf{\omega} \in (0, t]$ among the feature dimension, then use the cosine function to project all values to $\cos(t\mathbf{\omega}) \in [−1, +1]$. 
The selection of $\alpha, \beta$ depends on the scale of the maximum timestamp $t_{\max}$ we wish to encode. 
In order to distinguish all timestamps, we have to make sure $t_{\max} \times \alpha^{\frac{−(i -1)}{\beta}} \rightarrow 0$ as $i \rightarrow d$ to distinguish all timestamps. 
In practice, $d=100$ and $\alpha=\beta=\sqrt{d}$ works well for all datasets.
Notice that $\omega$ is fixed and will not be updated during training.
\begin{table}[ht]
    \centering
    \caption{Table of Notations}
    \footnotesize
    \begin{tabular}{lp{6.5cm}} 

        \toprule
        \textbf{Notation} & \textbf{Description} \\
        \midrule
        \( \mathcal{G} \) & Temporal network \\
        \( \mathcal{V} \) & Set of nodes of \( \mathcal{G} \) \\
        \( \mathcal{E} \) & Set of events of \( \mathcal{G} \) \\
        \( n \) & Number of nodes in \( \mathcal{G} \), n = \( |\mathcal{V}| \) \\
        \( m \) & Number of events in \( \mathcal{G} \), m = \( |\mathcal{E}| \)  \\
        \( G \) & Sequence of graph snapshots over time \\
        \( N \) & Number of time steps in snapshot sequence \( G \) \\
        \( G_k \) & Attributed snapshot graph at time step \( k \) \\
        \( E_k \) & Set of edges of \( G_k \) \\
        \( e_{ij} \) & Binary variable indicating the existence of an edge between nodes $v_i$ and $v_j$ \\        
        \( \mathbf{A}_k \) & Adjacency matrix of \( G_k \) \\
        \( \mathbf{X}_k \) & Node feature matrix of \( G_k \) \\
        \( \mathbf{E}_k \) & Edge feature matrix of \( G_k \) \\
        \( \mathbf{Z}_k \) &  Structural embedding matrix for \( G_k \) \\
        \( \mathbf{S}_k \) & Node state representation matrix of \( G_k \) \\
        \( \mathbf{s}_k \) & Graph state representation vector of \( G_k \) \\
        \( \mathbf{S}_k[i] \) & Representation vector of $i^{th}$ node in \( G_k \) \\
        \( \hat{\mathbf{Z}}^{(k)}_l \) & Local predictive encoding matrix at time step $k$ for future time step $l$ \\
        \( \hat{\mathbf{z}}^{(k)}_l \) & Global predictive encoding vector at time step $k$ for future time step $l$ \\        
        \( \hat{\mathbf{A}}_k \) & Reconstructed adjacency matrix for time step \( k \) \\
        \( \tilde{\mathbf{A}}_{k+1} \) & Predicted adjacency matrix for time step \( k+1 \) \\
        \( \hat{Z} \) & Set of local predictive encodings  \\
        \( \hat{z} \) & Set of global predictive encodings \\
        \( \hat{A} \) & set of reconstructed adjacency matrices \\
        \( \tilde{A} \) & set of predicted adjacency matrices \\
        \( l \) & Dummy index for time  \\
        \( L \) &  teneNCE objective function  \\
        \( L_{\text{pred}} \) & Prediction loss function \\
        \( L_{\text{recons}} \) & Reconstruction loss function \\
        \( L_{\text{cpc}} \) & Contrastive predictive coding (CPC) loss function\\
        \( \alpha \) & Weight hyperparameter for \( L_{\text{recons}} \) \\
        \( \beta \) & Weight hyperparameter for \( L_{\text{cpc}} \) \\
        \( P_{\text{neg}}(i, k) \) & Probability distribution of local negative sample indices for node \( i \) at time \( k \) \\
        \( P_{\text{neg}}(l) \) & Probability distribution of global negative sample indices at time step \( l \) \\
        \( \text{neg}_{(i, l)} \) & Set of local negative samples for node \( i \) at time \( l \) \\
        \( \text{neg}_{(l)} \) & Set of global negative samples at time \( l \) \\
        \( C \) & Temporal correlation coefficients \\
        \( C_i \) & Temporal correlation coefficient for node \( i \) \\
        \( d_\mathcal{V} \) & Dimension of node feature vector \\
        \( d_\mathcal{E} \) & Dimension of edge feature vector \\
        \( d_\textrm{enc} \) & Dimension of encoded node feature vector \\
        \( d_\textrm{state} \) & Dimension of node representation vector \\
        \bottomrule
    \end{tabular}
    \label{table:notations}
\end{table}
\subsection{Contrastive Predictive Coding}\label{subsec:infoNCE}
The original contrastive predictive coding technique was introduced in the well-known work \cite{oord2018representation} for learning context representation of a sequential data set $Y = \{y_1, \dots, y_M\}$.
The authors defined the infoNCE loss function to maximize the mutual information of the context representation $c_t$ with the future features of the input data $\{y_{t+1}, \dots, y_{M}\}$, where $t=1, \dots, M$.

Given a set of $M$ samples $Y = \{y_1, \dots, y_m\}$, with one positive sample drawn from the conditional distribution $p(y_{t+k}|c_t)$ and $M-1$ negative samples drawn from the distribution $p(y_{t+k})$, the loss function is defined as:
\begin{equation}
	L_{\textrm{M}} = -\mathbb{E}_M [\log \frac{f_k(y_{t+k}, c_t)}{\sum_{y_j \in Y} f_k(y_j, c_t)}],
\end{equation}
where $f_k(y_{t+k}, c_t) = \exp(z_{t+k}^\top W_k c_t)$ assigns a real positive score to the pair of context representation $c_t$ and future feature embedding $z_{t+k}$, using a bilinear map followed by the exponential function.

The infoNCE loss function can be viewed as the categorical cross-entropy loss of classifying the correct sample drawn from $p(y_{t+k}|c_t)$ from negative samples drawn from $p(y_{t+k})$.
The authors show that minimizing the infoNCE loss is equivalent to maximizing the mutual information between the context representation and future embeddings.

\subsection{Link Prediction Evaluation Metrics}\label{subsec:eval_metrics}

\textbf{Average Precision (AP score)}: It measures the precision of a binary classification model across multiple recall decision thresholds. The AP score is determined by:
    \begin{equation}
        \begin{aligned}
            \textrm{AP}_{\mathbf{A}_{k+1}}= \sum_b (\textrm{Rec}_b - \textrm{Rec}_{b-1}) \textrm{Prec}_b
        \end{aligned}
    \end{equation}
    Where $\textrm{Rec}_b$ and $\textrm{Prec}_b$ are the recall and precision at the $b^{th}$ decision threshold, according to Eq. \ref{eq:precision_recall}. $\textrm{Rec}_0 = 0$, $\textrm{Prec}_0 = 1$ and $\mathbb{I}$ denotes an indicator function.

    \begin{equation}
    \begin{aligned}
        \textrm{Rec}_{\mathbf{A}_{k+1}} &= \frac{\sum_{i,j} \mathbb{I}(\tilde{\mathbf{A}}_{k+1}[i,j] = 1) \cdot \mathbb{I}(\mathbf{A}_{k+1}[i,j] = 1)}{\sum_{i,j} \mathbb{I}(\mathbf{A}_{k+1}[i,j] = 1)}, \\
        \textrm{Prec}_{\mathbf{A}_{k+1}} &= \frac{\sum_{i,j} \mathbb{I}(\tilde{\mathbf{A}}_{k+1}[i,j] = 1) \cdot \mathbb{I}(\mathbf{A}_{k+1}[i,j] = 1)}{\sum_{i,j} \mathbb{I}(\tilde{\mathbf{A}}_{k+1}[i,j] = 1)}
    \end{aligned}
    \label{eq:precision_recall}
    \end{equation}

\textbf{AUC score}: This metric is calculated as the area under the receiver operating characteristic (ROC) curve. The ROC curve plots the true positive rate (TPR) against the false positive rate (FPR), and the area under this curve is a single scalar summary of the classifier’s performance. It is computed as follows:
    \begin{equation}
        \textrm{AUC}=\int_0^1 \textrm{TPR}\left(\textrm{FPR}^{-1}(t)\right) dt
    \end{equation}
    where the $\textrm{TPR}$ is calculated similarly to $\textrm{Rec}$ in Eq. \ref{eq:precision_recall} and the $\textrm{FPR}$ is given by:
    \begin{equation}
    \begin{aligned}
        \textrm{FPR}_{\mathbf{A}_{k+1}} = \frac{\sum_{i,j} \mathbb{I}(\tilde{\mathbf{A}}_{k+1}[i, j] = 1) \cdot \mathbb{I}(\mathbf{A}_{k+1}[i, j] = 0)}{\sum_{i,j} \mathbb{I}(\mathbf{A}_{k+1}[i, j] = 0)}
    \end{aligned}
    \end{equation}

\textbf{Mean Reciprocal Rank (MRR)}: This metric evaluates the ranking performance of a model by averaging the reciprocal ranks of the correct items in the predicted score list:
        \begin{equation}
    \textrm{MRR}_{\mathbf{A}_{k+1}} = \frac{1}{|{E}_{k+1}|} \sum_{(i, j) \in {E}_{k+1}} \frac{1}{\textrm{rank}_{ij}}
\end{equation}
        
    where $\textrm{rank}_{ij}$ is the rank of the true link $\mathbf{e}_{ij}$ in the sorted list of predicted scores.

\end{document}